\documentclass[conference]{IEEEtran}
\IEEEoverridecommandlockouts

\usepackage{cite}
\usepackage{amsmath,amssymb,amsfonts}
\usepackage{algorithmic}
\usepackage{graphicx}
\usepackage{textcomp}
\usepackage{xcolor}
\usepackage{url}

\usepackage[
    colorlinks=true,
    citecolor=blue,
    linkcolor=blue,
    urlcolor=blue
]{hyperref}

\usepackage{fancyhdr}
\def\BibTeX{{\rm B\kern-.05em{\sc i\kern-.025em b}\kern-.08em
    T\kern-.1667em\lower.7ex\hbox{E}\kern-.125emX}}

\begin{document}

\title{SAGE: Stability-Aware Graph-Based Ensemble Feature Selection for Explainable Postpartum Depression Risk Prediction}
\def\mycopyrightnotice{
  {\footnotesize 979-8-3195-2149-1/26/$31.00 ~\copyright2026 ~IEEE\hfill} % <--- Change here
  \gdef\mycopyrightnotice{}
}
\author{
\IEEEauthorblockN{Md. Rokon Islam Emon}
\IEEEauthorblockA{
\textit{Department of Computer Science}\\
\textit{Brunel University London}\\
Uxbridge UB8 3PH, United Kingdom\\
\texttt{2565915@brunel.ac.uk}
}
\and
\IEEEauthorblockN{Syed Shariar Alam Shuvo}
\IEEEauthorblockA{
\textit{Department of Law and Economics} \\
\textit{Technical University of Darmstadt} \\
Darmstadt, Germany\\
\texttt{shuvo.ss.121@gmail.com}
}
\and
\IEEEauthorblockN{Shahriar Siddique Ayon}
\IEEEauthorblockA{
\textit{Department of Computer Science}\\
\textit{American International University-Bangladesh}\\
Dhaka, Bangladesh\\
\texttt{shahriarayon63@gmail.com}
}
\and
\IEEEauthorblockN{Abdullah Al Mamun}
\IEEEauthorblockA{
\textit{Department of Computer Science and Engineering}\\
\textit{Dhaka University of Engineering and Technology}\\
Gazipur, Bangladesh\\
\texttt{mamun.duet.bd@gmail.com}
}
\and
\IEEEauthorblockN{Ahnaf Atef Choudhury}
\IEEEauthorblockA{
\textit{Department of Information Sciences and Technology}\\
\textit{George Mason University}\\
Fairfax, VA, USA\\
\texttt{achoudh9@gmu.edu}\\
% \textsuperscript{*}Corresponding author
}
}

\maketitle

%%%%%%%%%%%%%%%% Needs to be added for header information
% \thispagestyle{firstpage}
%%%%%%%%%%%%%%%%%%%%%%%%%%%%%%%%%%%%%

\begin{abstract}
Postpartum depression (PPD) poses a major burden on maternal and child health, especially in low- and middle-income countries where prevalence exceeds 19\%. Despite advancements in machine learning for PPD prediction, current approaches are limited by opaque global explanations that lack clinical usefulness at the patient level, unstable feature selection, and poor generalization under class imbalance. We propose SAGE, a Stability-Aware Graph-Based Ensemble feature selection system that incorporates both local explainable AI and a genetically optimized artificial neural network (GA-ANN). Using a primary cohort of 766 postpartum women, SAGE combines information-theoretic relevance, PCA-based structure, and graph-based interactions with bootstrap stability weighting to identify robust and non-redundant predictors. The GA-ANN architecture, optimized using a genetic algorithm and enhanced with GAN based oversampling, achieved strong performance with 87.96\% accuracy, 86.32\% F1 score, and 0.88 AUC using only 16 features, outperforming baseline and other feature selection methods. Psychological and socioeconomic factors such as EPDS score, PHQ-9 score, feelings about motherhood, and abuse history are the main predictors, while demographic factors have less influence. The LIME-based explanations allow instance-based insight into selected features from the graph, enabling personalized risk assessment. The findings make SAGE a scalable, interpretable, and clinical tool for early identification of PPD in health-care limited resources.
\end{abstract}


\section{Introduction}
Postpartum depression (PPD) is a common and serious condition after childbirth that affects mothers and can have lasting effects on families. The global prevalence of PPD is estimated at 17.22\%, with a disproportionate burden in low- and middle-income countries (19\%) compared to high-income countries (13\%) \cite{amer2024exploring}. According to the World Health Organization, over 20\% of mothers in developing countries suffer from PPD, which is associated with a higher risk of suicide and bad child outcomes such as poor growth, developmental delays, and illness \cite{WHO_maternal}. A meta-analysis of 86 studies found a 22.1\% prevalence of perinatal depression in rural areas and 21.1\% for postnatal depression, with higher rates of 25.4\% in Bangladesh due to economic hardship and limited mental health services, emphasizing the importance of early detection in South Asian resource-limited settings \cite{pan2024global}.

PPD arises from combined social, economic, obstetric, and psychological factors; low social support, marital dissatisfaction, and low income ($\approx 13\%$ higher risk) significantly increase vulnerability \cite{yang2026clinical}. Obstetric complications further worsen outcomes, with higher depression rates in affected mothers (23.97\%) compared to uncomplicated deliveries (16.71\%) \cite{wang2021mapping}. Given this complexity, artificial intelligence (AI) is essential in helping convert vast amounts of maternal healthcare data into effective clinical insights. Several recent studies have demonstrated strong performance using ensemble learning and optimized gradient boosting models, including XGBoost, for integrating obstetric and psychological factors with explainable AI capabilities \cite{huang2025postpartum,alam2025explainable}. These advances highlight the potential of AI for early identification of mothers at risk of PPD.

Many machine learning (ML)-based PPD studies achieve high accuracy; however, their reliability is often limited by weak validation, class imbalance, and unstable feature selection. Existing graph-based selectors may overlook statistical relevance and structural variability, while ensemble approaches often lack stability guarantees, leading to inconsistent feature rankings across resamples \cite{b8, mathew2025graph, harshitha2025emocare}. These limitations reduce interpretability and clinical trust, preventing current methods from achieving reliable real-world deployment.

To address these limitations, this study proposes a Stability-Aware Graph-based Ensemble (SAGE) feature selection framework that is coupled with a genetically optimized artificial neural networks (GA-ANN) and local explainable AI (XAI) for robust PPD prediction. SAGE combines information-theoretic, PCA, graph-based, and stability-weighted feature selection. Combined with GAN oversampling, it improves GA-ANN performance, while LIME provides patient-level explanations. Key Contributions of this work:

\begin{itemize}
    \item A novel stability-aware graph ensemble feature selector that explicitly models feature interactions and selection robustness.
    
    \item Analysis of a primary dataset of 766 postpartum women in Bangladesh shows that psychological and socioeconomic factors are the key predictors, while demographic factors are less influential.
    
    \item A GA-optimized ANN designed to enhance performance on imbalanced perinatal data using GAN-based oversampling.
    
    \item The first integration of local XAI with graph-based feature selection for patient-level PPD risk stratification.
    
\end{itemize}

The paper is organized as follows: Section \ref{sec:Lr} reviews related work, Section \ref{sec:Method} presents the proposed methodology, Section \ref{sec:Result} discusses the results, and Section \ref{sec:Con} concludes the study with future directions.

\section{Related Work}
\label{sec:Lr}

Recent studies indicate that AI can assist in predicting the risk of antenatal and PPD; however, its performance varies across different data sources. Clapp et al. used electronic health records (EHR) of 29,168 cases and found that prepartum Edinburgh Postnatal Depression Scale (EPDS) scores significantly improve early prediction of PPD \cite{b1}. Wang et al. used random forests to determine trends in PPD between 30.9\% and 29.1\% in 3,174 women, and found birth weight and maternal weight to be important indicators \cite{b2}. Another study, which used recursive feature elimination on 8,454 cases, achieved an AUC of 0.91 for predicting post-partum care \cite{b3}.

In studies on cesarean delivery data, XGBoost and neural network ensembles perform very well (AUC of 0.789-0.955, accuracy of 95.0\%) and are even better with hyperparameter tuning \cite{b4}. But results on cesarean delivery and other postpartum data sets are not always the same. For instance, there was an increase in performance (AUC 0.897-0.733 on the original data) after using SMOTE \cite{b6}. A study of 87 papers on PPD also reported that the average accuracy was high (nearly 93.4\%) \cite{alkhateeb2026ai}, but it was not very rigorous either, with less than half of the papers using internal or external validation.

%Prediction of PPD has changed from traditional models to deep learning (DL), but improvements in performance often make it harder to understand and evaluate in a clinical setting. Research on cesarean delivery data shows that models like XGBoost and neural network ensembles do very well (AUC up to 0.789–0.955, accuracy up to 95.0\%), and they often do even better when hyperparameters are tuned \cite{b4, b5}. However, the results from cesarean delivery and related postpartum datasets are not always consistent. For example, SMOTE caused performance to go up (AUC dropped from 0.897 to 0.733 on the original data) \cite{b6}. A review of 87 studies on PPD also found that the average accuracy was high (about 93.4\%) \cite{alkhateeb2026ai}, but the validation was not very strict, with fewer than half using proper internal or external validation.

Zhang et al. used LASSO and variance inflation factor analysis to identify 17 key predictors from 78 variables, achieving an AUC of 0.849 with XGBoost \cite{b7}. Cellini et al. combined LASSO and Boruta feature selection. Identified 11 important predictors, including gestational weight gain, in-law relationship and sleep quality for gradient boosting \cite{b8}. A systematic review highlighted that most studies rely on sociodemographic or clinical filtering without validating stability, which limits how well the models can be applied generally \cite{b9}.

%Dimensionality reduction is essential in PPD modeling to mitigate collinearity and overfitting in high-dimensional datasets. Zhang et al. utilized LASSO in conjunction with variance inflation factor analysis on 78 variables, identifying 17 significant predictors (e.g., antepartum depression, TSH, fetal weight), resulting in an AUC of 0.849 with XGBoost \cite{b7}. Huang et al. integrated LASSO and Boruta feature selection methodologies, utilizing their intersection to identify 11 pivotal predictors (e.g., gestational weight gain, in-law relationship, sleep quality) for gradient boosting \cite{b8}. A systematic review underscored that the majority of studies depend on simple sociodemographic or clinical filtering without stability validation, limiting model generalizability \cite{b9}.

Recent studies have expanded PPD prediction through diverse data modalities and validation strategies. Qi et al. demonstrated robust performance using externally validated ML models on large-scale clinical data \cite{qi2025prediction}. Wang et al. showed that plasma proteomics-based ML can provide objective, pre-symptomatic risk stratification \cite{wang2025plasma}. As predictive models move toward clinical use, obstetric decision-support systems should provide interpretable predictions. SHAP-based methods have been widely used to interpret PPD prediction models, including XGBoost, by identifying the contributions of key factors such as anxiety, pelvic floor muscle strength, and postnatal care satisfaction \cite{b8}. SHAP is also utilized for validating clinical cut-offs (e.g., 21.5\%) and making decisions regarding interventions among high-risk populations \cite{b4}. Most work still focuses on SHAP, while methods like LIME and counterfactual explanations are less commonly used.

%As predictive models get closer to being used in real life, obstetric decision-support systems need to be able to explain themselves. SHAP is commonly employed in PPD research to elucidate models such as XGBoost by detailing the influence of variables including pelvic-floor strength, anxiety, and satisfaction with postpartum care \cite{b8}. It has also been used to confirm clinical risk thresholds (like 21.5\%) and help make decisions about interventions in groups with a lot of risk \cite{b4}. Some studies combine SHAP with partial dependence plots to capture nonlinear effects, but this assumes feature independence, which is often unrealistic in obstetric data \cite{b7}. Most work still focuses on SHAP, while methods like LIME and counterfactual explanations are less commonly used.

However, current research often use weak validation, handle class imbalance poorly, and rely on unstable feature selection that ignores feature interactions and underutilizes rich postpartum features. This calls for the development of an improved and graph-oriented feature selection model with optimized models and local explanations. Applied to the NAFLD dataset with a SAGE-based Genetic Algorithm (GA) optimized ANN and GAN oversampling, the framework improved results from 98.35\% accuracy and 98.64\% F1-score to 98.46\% accuracy and 98.73\% F1-score, showing its effectiveness and novelty \cite{hossain2026explainable}. The following sections present the detailed methodology and results of our study.

%Existing studies often suffer from weak validation, poor handling of imbalanced data, and unstable feature selection methods that miss interactions and relationships between variables. They also rely mainly on global explanations, making it difficult to understand individual patient-level predictions. These gaps highlight the need for a stable, graph-aware feature selection framework with optimized models and local explainability for robust prediction.

\section{Methodology}
\label{sec:Method}
\begin{figure}
    \centering
    \includegraphics[width=0.99\linewidth]{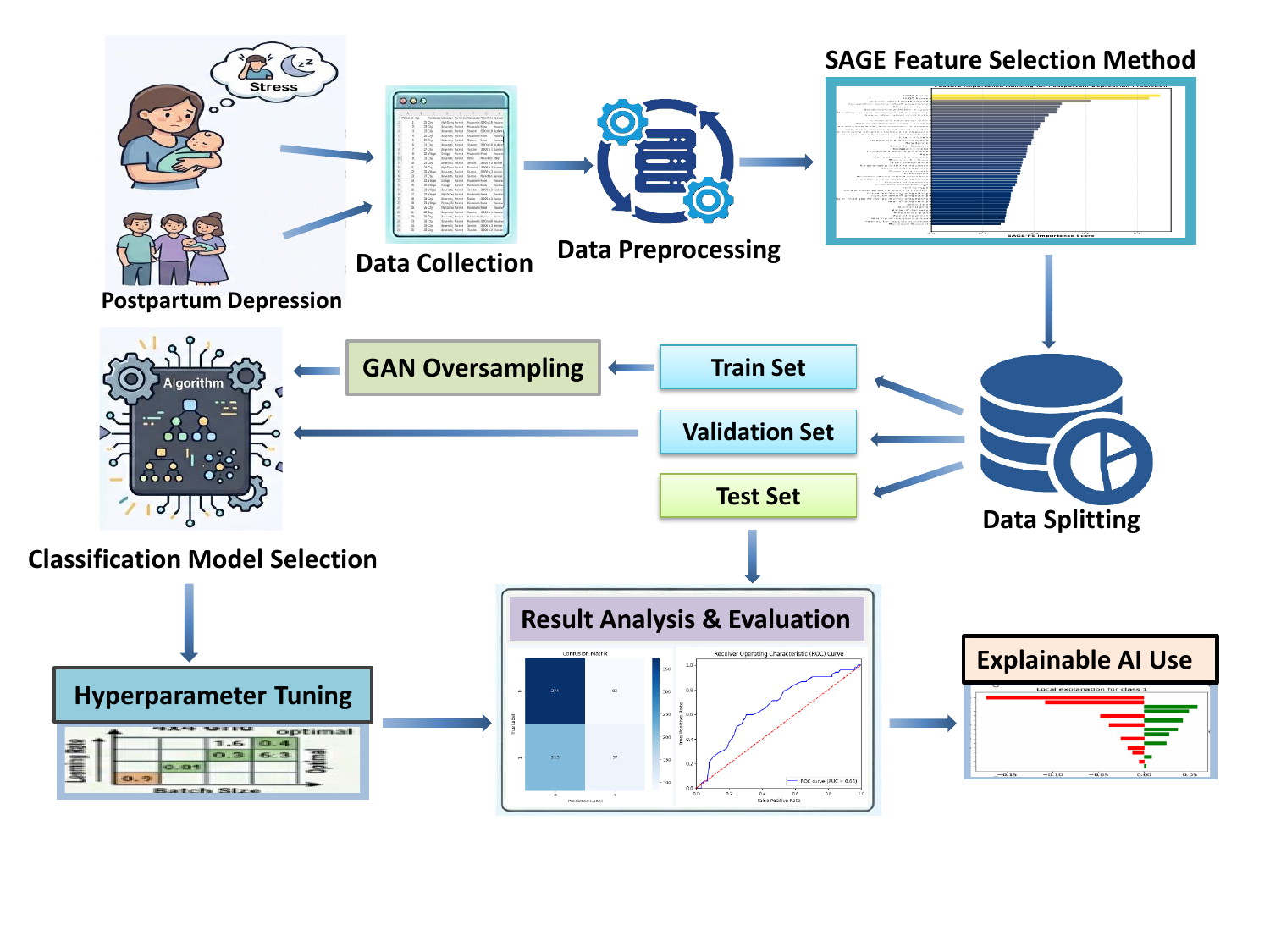}
    \caption{Proposed Framework for Postpartum Depression Prediction.}
    \label{fig:proposed}
\end{figure}

This section provides an overview of the proposed framework, data preparation, feature engineering, and development of an optimised model for women who have given birth. Data imbalance is addressed through oversampling and XAI is used for risk and protective factor identification. The proposed prediction framework is illustrated in Figure \ref{fig:proposed}.

%This section outlines the overall framework, including data preprocessing, feature selection, and optimized model development for postpartum women. Class imbalance is handled using oversampling, and XAI is used to identify key risk and protective factors. The proposed prediction framework is shown in Fig. \ref{fig:proposed}.

\subsection{Data Collection and Preprocessing}
The dataset used in this study is a primary field-collected dataset of 766 postpartum women in Bangladesh, gathered from hospital settings between March and June 2025 across urban, rural, and outpatient facilities \footnote{Available: \url{https://data.mendeley.com/datasets/nzgnsrgsg5/1}}. Participants aged 18–41 were recruited within 24 months of childbirth following ethical guidelines, ensuring informed consent and anonymity. Mental health status was assessed using standardized instruments, including Patient Health Questionnaire (PHQ)-2, PHQ-9, and EPDS, with 40.47\% of participants screening positive for depression during pregnancy. The dataset includes diverse sociodemographic, clinical, obstetric, psychosocial, and neonatal factors, covering maternal characteristics, pregnancy history, delivery details, health conditions, and postnatal behaviors. 

%This extensive and multifaceted dataset facilitates predictive modeling, risk factor analysis, and interpretable AI-driven analysis for postpartum depression.

At first, the dataset had 51 features and 766 patient records. Irrelevant identifiers such as Patient ID were removed, along with redundant outcome columns (PHQ-9 Result and EPDS Result) since their corresponding score values were retained. There are no missing values in the dataset. Instead of being nulls, entries marked as "None" were treated as valid categorical values. There are four numerical features, and the other categorical variables were turned into numerical data using label encoding. The final dataset has 48 features and 766 instances after preprocessing.

\subsection{Stability-Aware Graph-Based Ensemble Feature Selection}
After preprocessing, redundant features were reduced using Recursive Feature Elimination (RFE), Tree-based Feature Importance, Principal Component Analysis–Information Gain (PCA-IG), and the proposed Stability-Aware Graph Ensemble (SAGE) method. SAGE feature selection combines statistical relevance, structural variance, and graph-based interaction information in a single framework, which is weighted to account for stability in feature selection for classification problems. Information gain and PCA-based structural contribution are defined as: 

\begin{equation}
IG_j = I(X_j; y), \quad PCA_j = \sum_{k=1}^{m} |V_{jk}| \cdot \lambda_k
\end{equation}

While graph-based interaction importance and final ensemble scoring are given by:

\begin{equation}
C_j = \frac{\mathrm{deg}(X_j)}{d - 1}, \quad
Score_j = \alpha IG_j^{*} + \beta PCA_j^{*} + \gamma C_j^{*}
\end{equation}

Where stability-adjusted components $IG_j^*$, $PCA_j^*$, and $C_j^*$ are derived using bootstrap statistics as:

\begin{equation}
S_j^{*} = \frac{\mu_j}{\sigma_j + \epsilon}
\end{equation}

Here, $X_j$ and $y$ represent the $j$-th feature and target variable, respectively; $I(X_j;y)$ denotes mutual information, $V_{jk}$ and $\lambda_k$ are PCA loadings and explained variance, $\deg(X_j)$ represents graph degree, and $\mu_j$, $\sigma_j$ denote the mean and standard deviation of feature importance across $B$ bootstrap resamples. $\epsilon$ is a smoothing constant, while $\alpha$, $\beta$, and $\gamma$ are ensemble weights satisfying $\alpha+\beta+\gamma=1$.

The proposed feature selection framework combines IG, PCA, and graph centrality. An undirected feature graph is constructed from the Pearson correlation matrix, where an edge $(i,j)$ is formed when $|\rho_{ij}|\geq0.5$, and $C_j$ denotes the normalized degree centrality. For each component, $B=100$ bootstrap resamples are used to compute the stability-adjusted score $S_j^{*}=\mu_j/(\sigma_j+\epsilon)$, where $\epsilon=10^{-6}$. The component scores are min--max normalized to $[0,1]$ and combined using equal weights $\alpha=\beta=\gamma=1/3$, with sensitivity analysis confirming ranking robustness.

SAGE has a computational complexity of $O(B \cdot n \cdot d + d^3)$, dominated by bootstrap resampling and PCA. For our dataset ($n=766$, $d=48$), it completes in under one second on standard hardware and is implemented in Python using \texttt{scikit-learn} and \texttt{NetworkX}.

\begin{figure}[htbp]
    \centering
    \includegraphics[width=0.99\linewidth]{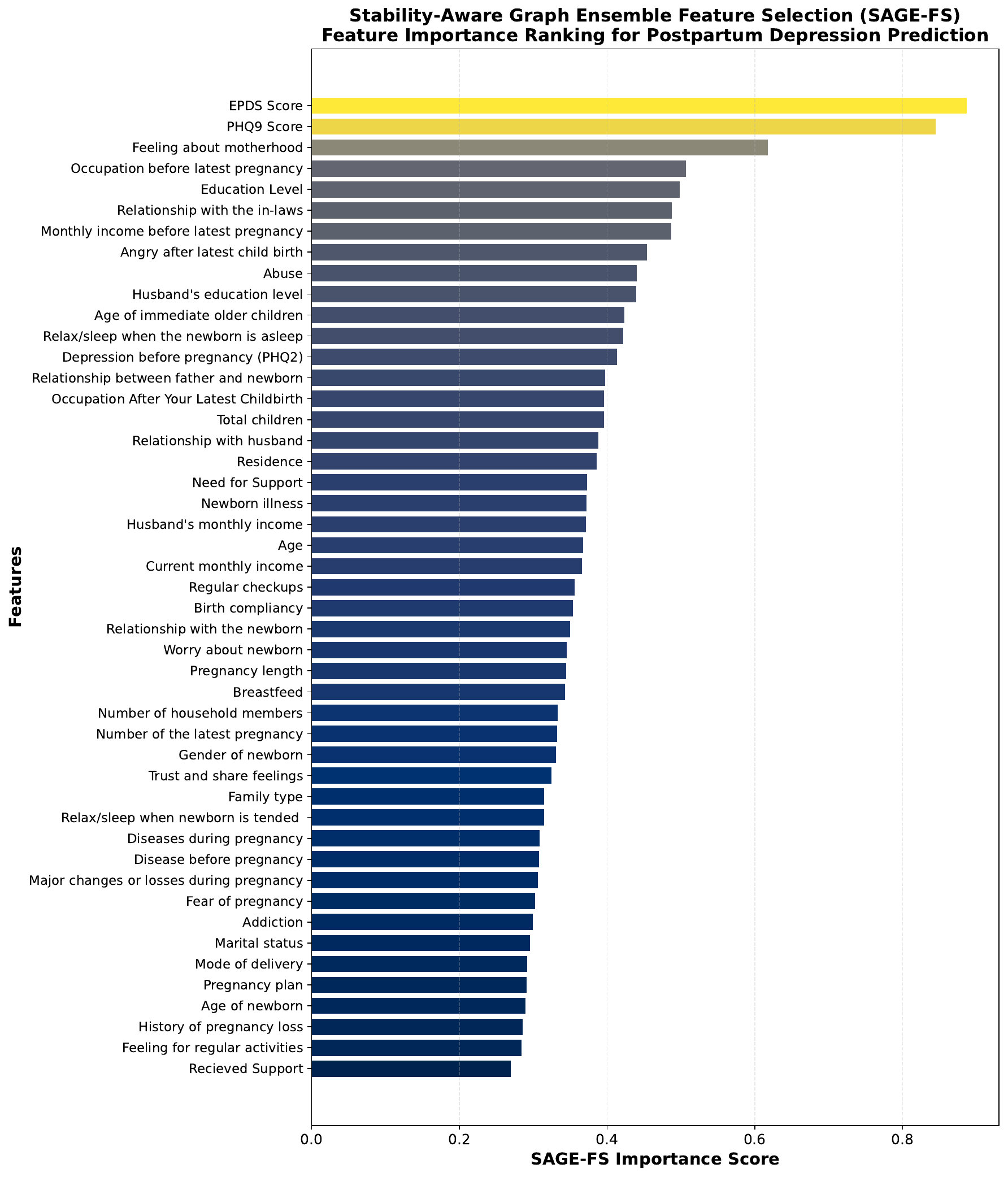}
    \caption{SAGE Feature Importance Ranking for Postpartum Depression Prediction.}
    \label{fig:feature1}
\end{figure}

Figure~\ref{fig:feature1} presents the SAGE-based feature importance ranking, where EPDS Score (0.89) and PHQ-9 Score (0.84) emerge as the most dominant predictors. Other key factors include feeling about motherhood (0.62), occupation (0.51), education (0.50), abuse (0.44), and anger after childbirth (0.45), reflecting the strong impact of psychological and socioeconomic variables. Moderate contributions are observed from family and economic factors such as relationship with in-laws and income ($\approx 0.49$), while variables like support received, pregnancy loss history, and mode of delivery show lower importance ($\approx 0.27$--$0.29$). Overall, psychological and emotional factors drive prediction, with demographic and clinical variables playing a supporting role.

Based on SAGE ranking, the least important features were removed, resulting in a final dataset of 17 features (including the target) with 766 observations. Stratified 5-fold cross-validation was used for unbiased evaluation, with each fold allocating 80\% for training and 20\% for testing; 20\% of the training portion was reserved for validation. SAGE feature selection, CTGAN oversampling, and GA-ANN tuning were performed exclusively on each training split, with the resulting features and hyperparameters applied unchanged to the validation and test sets. Final performance was averaged across the five test folds.

\subsection{Model Selection and Hyperparameter Tuning}
Using standard Python libraries, we evaluated multiple ML and DL models with optimized hyperparameters to improve prediction performance. Baseline models include Logistic Regression (LR), Support Vector Classifier (SVC), K-Nearest Neighbors (KNN), and Random Forest (RF), all optimized via grid search. An Artificial Neural Network (ANN) with two hidden layers (64, 32 neurons), ReLU activation, dropout (0.3), and Adam optimizer (lr=0.001) was also used for binary classification. ANN outperformed all baseline models on the full feature set, and its performance was further improved through Genetic Algorithm (GA) optimization. The GA optimizes the ANN using 20 individuals over 50 generations, with tournament selection (size 3), single-point crossover ($P_c=0.8$), and Gaussian mutation ($P_m=0.2$). The search covers $n\in\{16,32,64,128\}$, $\eta\in[10^{-4},10^{-2}]$, $d\in[0.1,0.5]$, $b\in\{16,32,64\}$, and $e\in\{50,100,200\}$, using mean 5-fold CV F1-score as the fitness function. 

%In this study, the search space was defined as:

%\begin{equation}
%n \in [32,128], \quad 
%\eta \in [10^{-4},10^{-2}], \quad 
%d \in [0.2,0.5], \quad 
%b \in [16,64], \quad 
%e \in [20,100]
%\end{equation}

%Fitness is evaluated as $Fitness(\theta) = Accuracy_{val}$ or $F1_{val}$, and the GA evolves the solution toward $\theta^{*} = \arg\max_{\theta \in \Omega} Fitness(ANN(\theta))$, resulting in improved accuracy and generalization performance.

The target variable 'Depression during pregnancy (PHQ-2)' is imbalanced, with 59.53\% negative and 40.47\% positive cases. To address class imbalance, Conditional Tabular GAN (CTGAN)-based oversampling was applied exclusively within each training fold, leaving the test set untouched for unbiased evaluation. CTGAN was trained for 300 epochs with $256\times128$ generator/discriminator hidden dimensions, batch size 500, dropout 0.3, and mode-specific normalization for categorical variables, with early stopping based on validation performance. Synthetic samples were generated until a 1:1 minority-to-majority ratio was achieved.

\subsection{Explainable AI Integration}
XAI improves interpretability by explaining predictions. This study uses LIME for instance-level explanations to identify case-specific feature contributions \cite{ayon2026explainable}. The LIME model is set up like this:

\begin{equation} 
\hat{g}\left( x \right)=arg min_{g\epsilon G} L\left( f,g,\pi_{x\acute{}} \right)+\Omega\left( g \right) 
\end{equation}

In LIME, the original complex model is locally approximated by an interpretable surrogate model $\hat{g}(x) \in G$, obtained by minimizing a loss function $L(f, g, \pi_{x'})$ that balances fidelity to the original model and interpretability.

%In LIME, the model is locally approximated with an interpretable surrogate $\hat{g}(x)$ by minimizing $L(f, g, \pi_{x'})$ to balance fidelity and interpretability.

\subsection{Evaluation Metrics}
The classification models were evaluated using standard performance metrics, including accuracy, precision, recall, F1-score, and the Area Under the Receiver Operating Characteristic Curve (AUC-ROC). These metrics were computed from the confusion matrix, which consists of true positives (TP), true negatives (TN), false positives (FP), and false negatives (FN). The AUC-ROC measures the model's ability to distinguish between classes across different decision thresholds, where values closer to 1 indicate better predictive performance.

\section{Results Analysis and Discussion}
\label{sec:Result}
All models were implemented in Python on Google Colab using TensorFlow 2.13, Scikit-learn 1.3, DEAP 1.4, and SDV 1.2 (CTGAN). Stratified 5-fold cross-validation with a fixed random seed of 42 was employed to ensure robust and reproducible evaluation. This section reports the best model, comparisons before and after feature selection, effects of GAN-based imbalance handling, and LIME-based feature explanations.

\subsection{Performance Evaluation of Baseline Models}
Table \ref{table:performance} shows model performance using the full feature set without oversampling. GA-ANN achieves the best results (79.28\% accuracy, 76.52\% F1-score, 0.79 AUC), followed by ANN (77.67\%, 74.48\%, 0.78) and RFC (75.58\%, 70.72\%, 0.76). Among baseline models, LR performs best (73.43\% accuracy), while SVC and KNN show lower performance, highlighting the superiority of GA-based deep learning approaches.

%Table \ref{table:performance} presents the comparative performance of different models using the full feature set without oversampling. The GA-ANN achieves the best performance with 79.28\% accuracy, 76.52\% F1-score, and 0.79 AUC, followed by ANN (77.67\%, 74.48\%, 0.78) and RFC (75.58\%, 70.72\%, 0.76). Among baseline models, LR performs best (73.43\% accuracy), while SVC and KNN show comparatively lower results, indicating the superiority of DL and GA-based optimization.

\begin{table}[htbp]
\centering
\caption{Model Performance with Full Feature Set and No Oversampling}
\begin{tabular}{p{1.2cm} p{1cm} p{0.9cm} p{0.9cm} p{1cm} p{0.9cm}}
\hline
\textbf{Model} & \textbf{Precision (\%)} & \textbf{Recall (\%)} & \textbf{F1-score (\%)} & \textbf{Accuracy (\%)} & \textbf{AUC} \\ \hline
LR & 69.31 & 65.34 & 68.61 & 73.43 & 0.74 \\ \hline
SVC & 66.87 & 64.78 & 65.04 & 71.54 & 0.72 \\ \hline
KNN & 66.35 & 64.23 & 64.82 & 71.21 & 0.71 \\ \hline
RFC & 71.36 & 69.67 & 70.72 & 75.58 & 0.76 \\ \hline
ANN & 75.12 & 73.35 & 74.48 & 77.67 & 0.78 \\ \hline
GA-ANN & 77.06 & 75.36 & 76.52 & 79.28 & 0.79 \\ \hline
\end{tabular}
\label{table:performance}
\end{table}

\subsection{Feature Selection and Oversampling for Performance Enhancement}
Table \ref{table:performance_oversampling} compares model performance under different feature selection methods with GAN-based oversampling. SAGE-based feature selection using 16 features achieves the best results, where GA-ANN attains 87.96\% accuracy, 86.32\% F1-score, and 0.88 AUC. PCA-IG with 15 features also performs strongly, with GA-ANN reaching 87.42\% accuracy and 0.87 AUC. RFE + Tree-based methods using 19 features show slightly lower but competitive performance, with GA-ANN achieving 86.18\% accuracy and 0.86 AUC. Overall, GA-ANN consistently outperforms other models, with SAGE providing the most effective feature representation.

\begin{table}[htbp]
\centering
\caption{Model Performance with Feature Selection and GAN Oversampling}
\resizebox{\columnwidth}{!}{%
\begin{tabular}{p{0.8cm}p{1.2cm}p{1cm}p{0.8cm}p{0.8cm}p{1cm}p{0.8cm}}
\hline
\textbf{FS Method} & \textbf{Model} & \textbf{Precision (\%)} & \textbf{Recall (\%)} & \textbf{F1-score (\%)} & \textbf{Accuracy (\%)} & \textbf{AUC} \\ \hline

RFE +                & ANN        & 85.12 & 84.94 & 85.02 & 85.52 & 0.86  \\
Tree                 & RFC        & 84.77 & 84.45 & 84.72 & 85.31 & 0.85  \\
Based                & GA-ANN     & 85.65 & 85.38 & 85.60 & 86.18 & 0.86  \\ \hline

                     & LR         & 84.62 & 84.08 & 84.47 & 85.63 & 0.86  \\
PCA-IG               & ANN        & 85.44 & 85.27 & 85.32 & 86.25 & 0.86  \\
                     & GA-ANN     & 86.23 & 85.74 & 86.17 & 87.42 & 0.87  \\ \hline

                     & RFC        & 84.81 & 83.95 & 84.67 & 85.76 & 0.86  \\
SAGE                 & ANN        & 85.18 & 85.04 & 85.10 & 86.43 & 0.87  \\
                     & \textbf{GA-ANN} & \textbf{86.46} & \textbf{85.56} & \textbf{86.32} & \textbf{87.96} & \textbf{0.88}  \\
\hline
\end{tabular}%
}
\label{table:performance_oversampling}
\end{table}

\begin{figure}[htbp]
    \centering
    \includegraphics[width=0.99\linewidth]{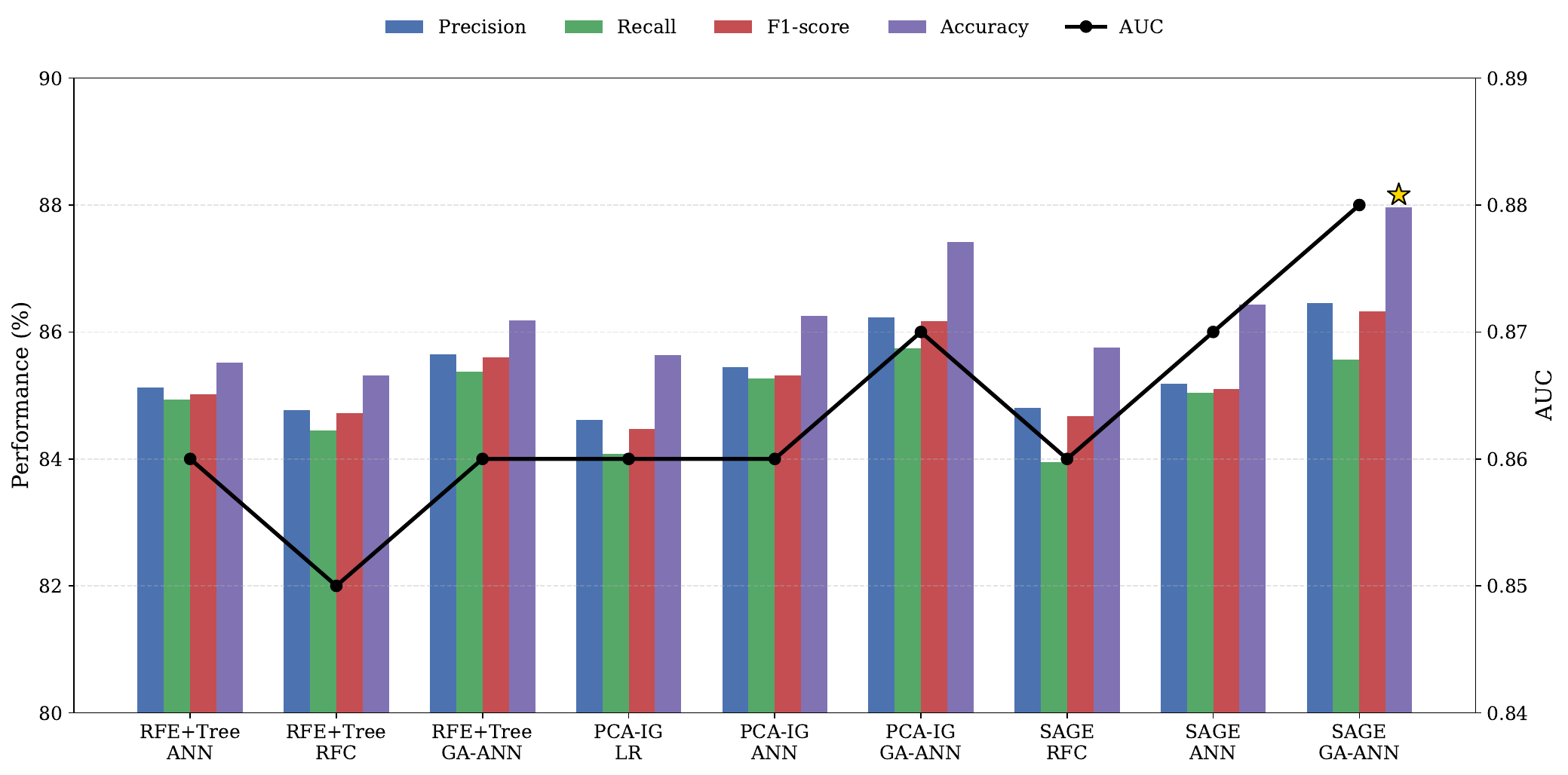}
    \caption{Comparison of Model Performance under Different Feature Selection Methods with GAN Oversampling.}
    \label{fig:fig04}
\end{figure}

Figure \ref{fig:fig04} compares the performance of different machine learning models using three feature selection methods (RFE+Tree, PCA-IG, and SAGE) after GAN-based oversampling. The evaluation includes precision, recall, F1-score, accuracy, and AUC. Overall, the SAGE feature selection method consistently achieved the strongest performance, with the GA-ANN model producing the highest results across all evaluation metrics, including 87.96\% accuracy and an AUC of 0.88. These findings demonstrate that combining SAGE feature selection with GA-ANN provides the most effective classification performance among the evaluated approaches.

\subsection{LIME-Based Insights into Model Decisions}
\begin{figure}[htbp]
    \centering
    \includegraphics[width=0.99\linewidth]{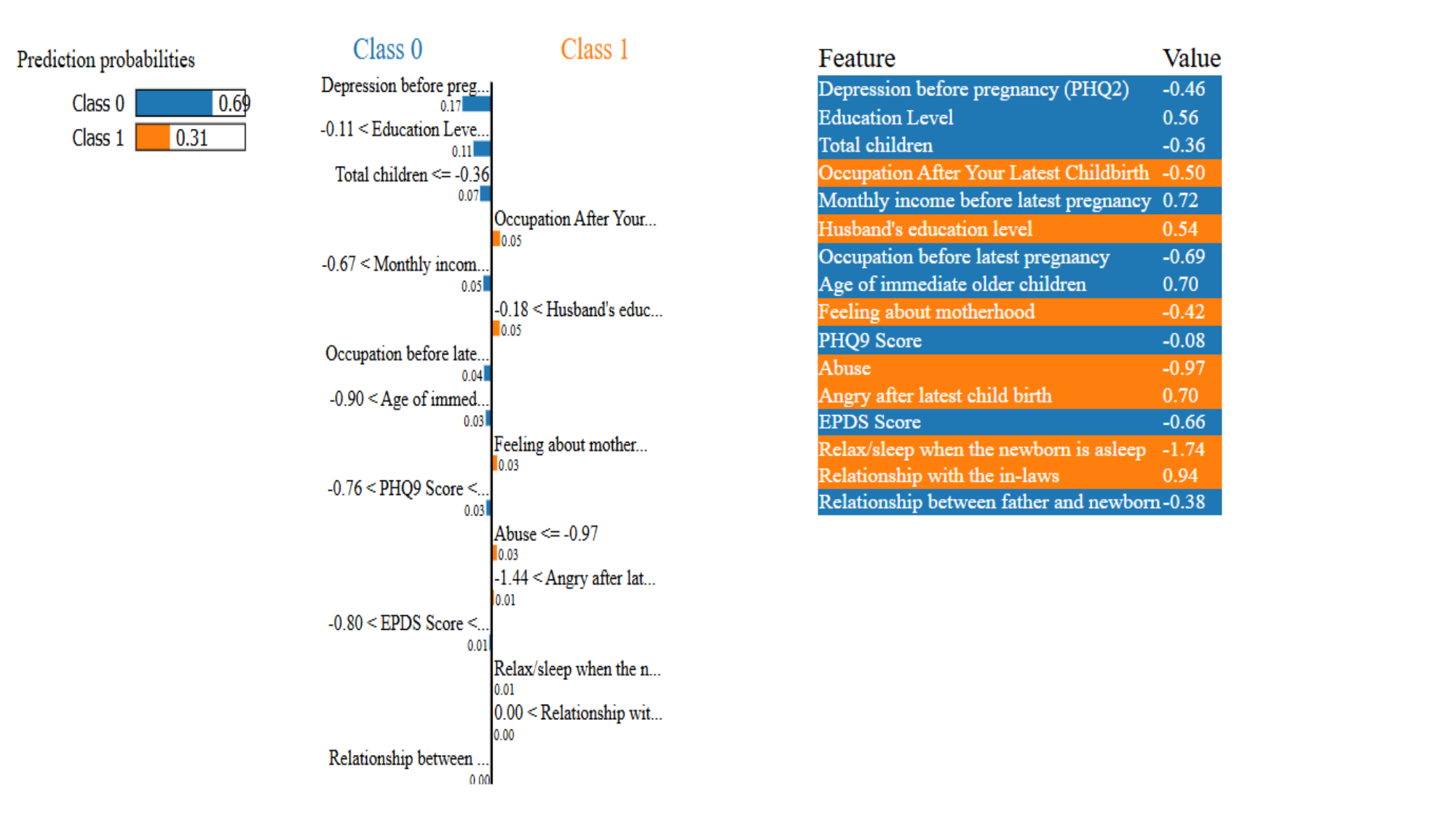}
    \caption{LIME-Based Local Explanation of Model Prediction.}
    \label{fig:fig3}
\end{figure}

The impact of features on predictions is shown using LIME tabular and feature importance plots based on SAGE-selected features and the GA-ANN model. Figure \ref{fig:fig3} shows a LIME-based explanation where the model predicts Class 0 with 69\% probability and Class 1 with 31\%. Key protective factors include relationship with in-laws (0.94), income (0.72), age of older children (0.70), education (0.56), and husband’s education (0.54), while risk factors include poor sleep (-1.74), abuse (-0.97), EPDS score (-0.66), occupation after childbirth (-0.50), and feeling about motherhood (-0.42). Overall, protective socioeconomic factors outweigh psychosocial stressors, leading to a final prediction of Class 0.

\begin{figure}[htbp]
    \centering
    \includegraphics[width=0.99\linewidth]{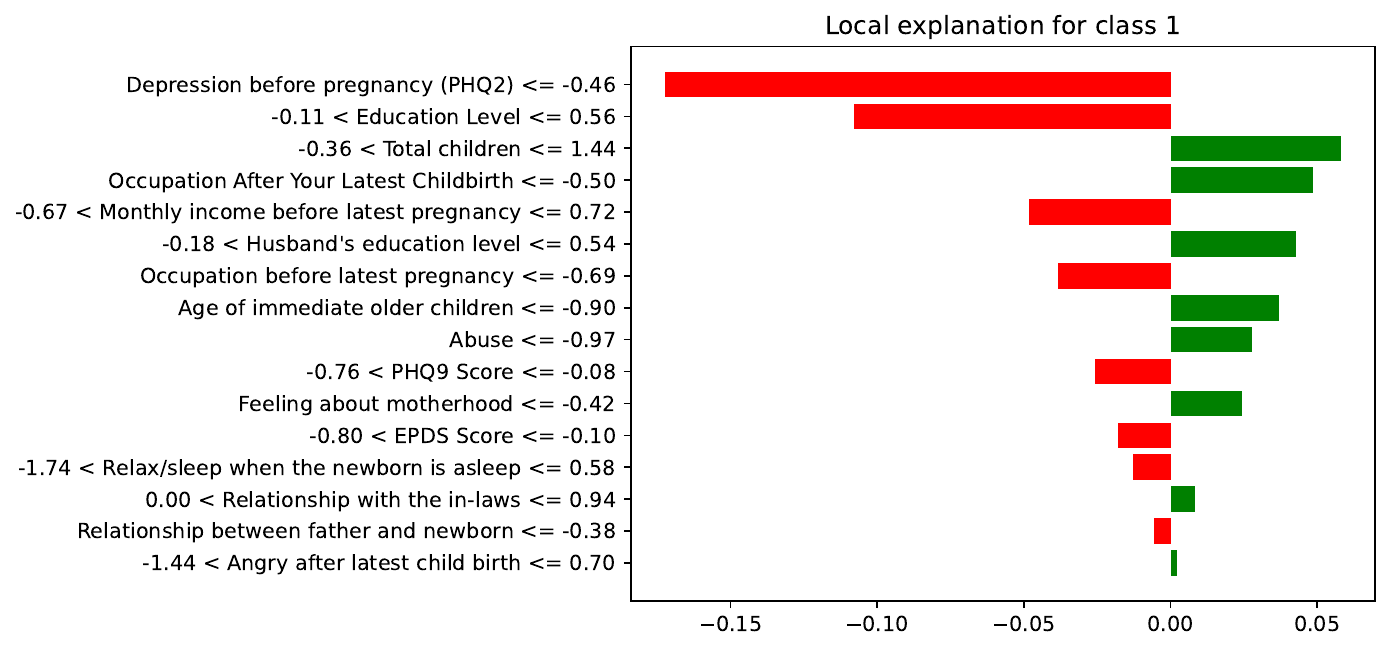}
    \caption{Instance-Level Feature Importance Using LIME.}
    \label{fig:fig4}
\end{figure}

Figure \ref{fig:fig4} presents the LIME feature importance plot for a Class~1 prediction, where key risk factors such as prior depression (PHQ2 $\approx -0.16$), education level ($\approx -0.11$), income, PHQ-9 ($\approx -0.03$), EPDS ($\approx -0.02$), and poor sleep ($\approx -0.02$) drive the model toward higher risk. In contrast, factors like total children ($\approx +0.06$), husband’s education ($\approx +0.04$), and post-childbirth occupation ($\approx +0.04$) contribute toward Class~0. Overall, negative psychological and lifestyle factors dominate, leading to a higher-risk prediction (Class 1).

Compared with existing studies, SAGE provides a strong balance of predictive performance, feature parsimony, stability, and interpretability. Zhang et al. \cite{b7} reported an AUC of 0.849 using 17 predictors, whereas SAGE achieves an AUC of 0.88 with only 16 features. Cellini et al. \cite{b8} selected 11 predictors using LASSO-Boruta but did not assess selection stability, which SAGE addresses through bootstrap-weighted ensemble selection. Although Liu et al. \cite{b4} reported up to 95.0\% accuracy on cesarean-specific cohorts, SAGE is evaluated on a more heterogeneous primary cohort. Moreover, SAGE combines graph-based feature selection with local LIME explanations, extending beyond the global SHAP-based interpretability used in prior studies \cite{b8, b4, sharif2024comparative}. SAGE demonstrates improved methodological rigor while maintaining competitive predictive performance and clinically relevant interpretability.

%Previous work is largely unidirectional, treating mental health as input with binary or coarse outputs, lacking class-specific SHAP interpretability and advanced imbalance handling or optimization for robust multiclass prediction \cite{chike2026interpretable, tang2026development, han2026adverse, allani2025interactive, kim2026explainable}. We propose a novel framework that addresses these gaps by comparing SHAP with traditional feature selection methods, where PCA-IG performs best among conventional approaches, but SHAP achieves superior performance. The framework also enables effective analysis of mental health risk factors by identifying the most influential features.

\section{Conclusion and Future Work}
\label{sec:Con}
This study presents SAGE, an ensemble feature selection framework for PPD prediction that incorporates a GA-optimized ANN and a local XAI. By combining multiple selection methods with stability weighting, the model identifies a reliable set of key features from high-dimensional data. The proposed system addresses the class imbalance and achieves better accuracy than traditional methods. Our findings indicate that psychological and sociodemographic factors are the main risk factors of PPD. Moreover, LIME explanations enhance the interpretability by providing patient-specific insights, which makes it suitable for clinical decision support for maternal health.

The framework should be tested across regions to assess whether similar patterns exist among women in South Asia and Africa. Longitudinal analysis of the prenatal and postnatal periods can improve the understanding of temporal risk. Multimodal data (EHR, wearable data and clinical text) can be used to enhance predictive performance, while stability-aware feature selection strategies can still be applied. The system should be extended to other perinatal mental health problems and clinical systems.

\section*{Data Availability}  
The dataset is available on Mendeley at the following link:
\url{https://data.mendeley.com/datasets/nzgnsrgsg5/1}

\bibliography{ref} % Assuming the file is named "references.bib"
\bibliographystyle{IEEEtran}

\end{document}